\documentclass[conference]{IEEEtran}
\IEEEoverridecommandlockouts
\usepackage{cite}
\usepackage{amsmath,amssymb,amsfonts}
\usepackage{algorithmic}
\usepackage{graphicx}
\usepackage{textcomp}
\usepackage{xcolor}
\usepackage{colortbl} % for cell colors
\usepackage{changepage,threeparttable} % for wide tables
\usepackage{soul}% for underlines
\usepackage{booktabs, multirow} % for borders and merged ranges
\usepackage{pifont} % for ✓ (check mark) and ✗ (ballot x)
\usepackage{pgfplots}
\pgfplotsset{compat=1.18}
\usepackage{arydshln} % 提供 \cdashline 命令
\usepackage{hyperref}

\def\BibTeX{{\rm B\kern-.05em{\sc i\kern-.025em b}\kern-.08em
    T\kern-.1667em\lower.7ex\hbox{E}\kern-.125emX}}

\title{CPJ: Explainable Agricultural Pest Diagnosis via Caption-Prompt-Judge with LLM-Judged Refinement}

\author{
\IEEEauthorblockN{
Wentao Zhang\IEEEauthorrefmark{1},
Tao Fang\IEEEauthorrefmark{2}\thanks{Corresponding Author: Tao Fang; Email: taofang@mmc.edu.mo}, % 标注第一位
Lina Lu\IEEEauthorrefmark{1}\thanks{Co-corresponding Author: Lina Lu; Email: 3050651@qq.com}, % 标注第二位
Lifei Wang\IEEEauthorrefmark{2},
Weihe Zhong\IEEEauthorrefmark{2}
}
\IEEEauthorblockA{\IEEEauthorrefmark{1}Business School, Shandong University of Technology, Shandong, China}
\IEEEauthorblockA{\IEEEauthorrefmark{2}Institute of International Language Services Studies, Macau Millennium College, Macau SAR, China}
}

\begin{document}

\maketitle

\begin{abstract}
Accurate and interpretable crop disease diagnosis is essential for agricultural decision-making, yet existing methods often rely on costly supervised fine-tuning and perform poorly under domain shifts. We propose Caption--Prompt--Judge (CPJ), a training-free few-shot framework that enhances Agri-Pest VQA through structured, interpretable image captions. CPJ employs large vision-language models to generate multi-angle captions, refined iteratively via an LLM-as-Judge module, which then inform a dual-answer VQA process for both recognition and management responses. Evaluated on CDDMBench, CPJ significantly improves performance: using GPT-5-mini captions, GPT-5-Nano achieves \textbf{+22.7} pp in disease classification and \textbf{+19.5} points in QA score over no-caption baselines. The framework provides transparent, evidence-based reasoning, advancing robust and explainable agricultural diagnosis without fine-tuning. Our code and data are publicly available.\footnote{\url{https://github.com/CPJ-Agricultural/CPJ-Agricultural-Diagnosis}}
\end{abstract}


\section{Introduction}
\label{sec:introduction}

Accurate and interpretable crop disease diagnosis is critical for agricultural decision-making. Conventional vision models (e.g., YOLO) only provide categorical labels without integrating essential context such as environment or crop variety, hindering practical utility as real-world diagnosis demands deeper reasoning combining visual evidence and expert knowledge. The opacity of such black-box systems undermines trust, emphasizing the need for explainable AI (XAI)~\cite{barredo2020explainable}.

Advanced large vision-language models (LVLMs), such as Qwen-VL and GPT-series, demonstrate strong performance in open-domain visual question answering (VQA)~\cite{zhang2024visual,li2023blip}. Yet when applied to Agri-Pest VQA, these models often rely on raw image data alone, resulting in incomplete interpretations that miss critical diagnostic details distinguishing pathogen types. Domain-specific approaches like ITLMLP for cucumber disease recognition~\cite{cao2023cucumber} and multimodal foundation models for agricultural applications~\cite{tan2023promises} show promising directions but lack comprehensive explainability frameworks.

Existing works, including multimodal datasets like AgMMU and CDDM~\cite{gauba2025agmmu,liu2024multimodal}, focus on fine-tuning~\cite{radford2021learning, alayrac2022flamingo} for accuracy but neglect structured explainability, leading to unreliable outputs under domain shifts~\cite{saadati2024out}. Without intermediate representations, models cannot connect visual cues to causal explanations, reducing interpretability and robustness.

We propose enhancing Agri-Pest VQA with explicit, interpretable image captions as auxiliary inputs. We employ a larger LVLM to generate multi-angle descriptive captions detailing plant types, disease symptoms, severity, and uncertainties. These captions act as transparent intermediates, augmenting the model's understanding like human experts verbalizing observations before diagnosis, thereby improving explainability through traceable, human-readable artifacts~\cite{wei2025lightweight}.

\begin{figure*}[!t]
    \vspace{-1em} % 缩小顶部间距
    \centering
    \includegraphics[width=0.96\textwidth]{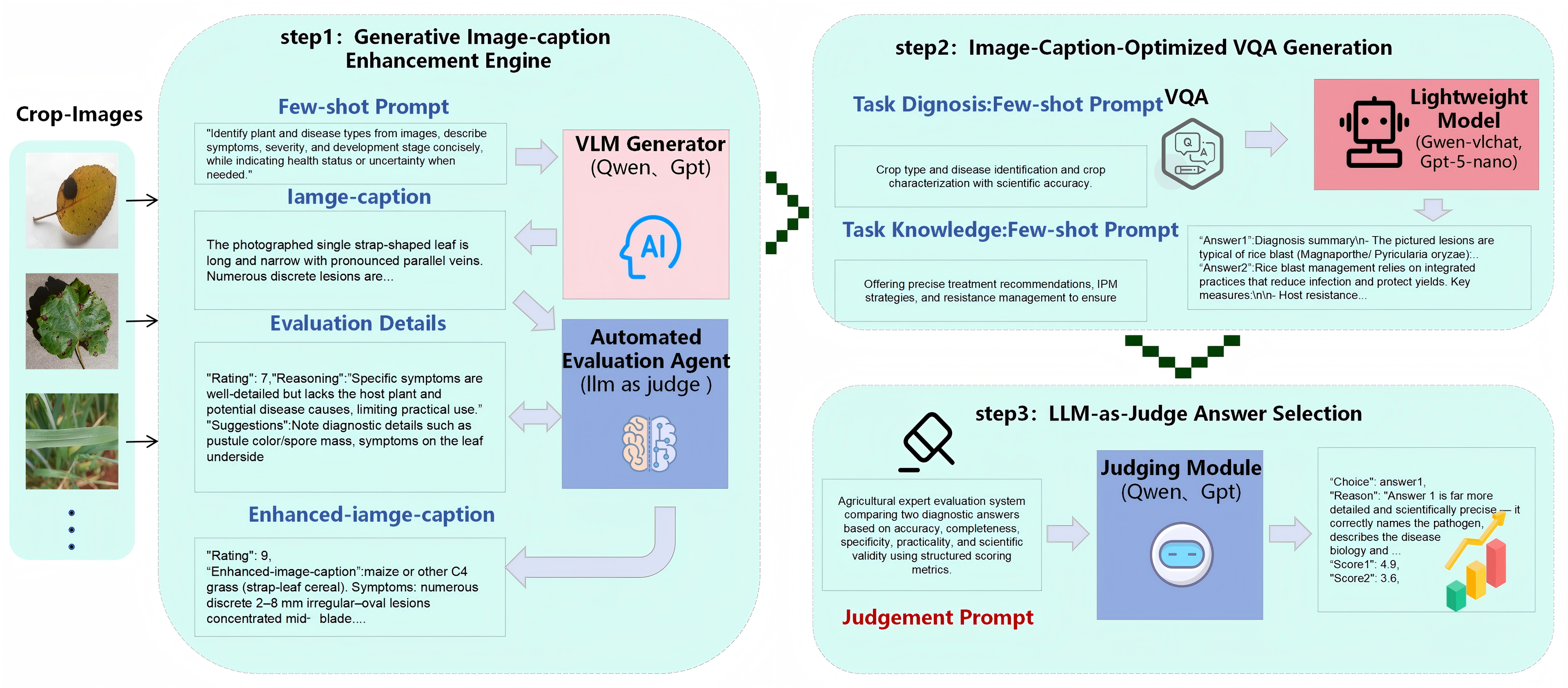} % 换成你的文件名
    \vspace{-4mm} % 缩小底部间距
    \caption{
    Overview of the ``CPJ" pipeline for explainable Agri-Pest VQA, featuring three cohesive stages: (i) \textbf{Generative Explanational Captioning}: A large vision-language model analyzes input crop disease images to generate multi-angle descriptive captions detailing plant morphology, disease symptoms, severity levels, and diagnostic uncertainties. These captions undergo iterative refinement via LLM-as-a-Judge evaluation to ensure accuracy, completeness, and neutrality. (ii) \textbf{Task-Specific Prompt-Based VQA Generation}: Using the refined captions alongside original images and few-shot exemplars, the VQA model generates dual complementary answers addressing both disease recognition (crop type, disease identification, visual features) and actionable management guidance (treatment methods, prevention strategies). (iii) \textbf{LLM-as-a-Judge Answer Selection}: A stronger LLM evaluates both candidate answers against reference answers using multi-dimensional criteria (factual accuracy, completeness, specificity, practical relevance), selecting the superior response to deliver reliable, interpretable diagnostic recommendations.
    }
    \label{fig:framework}
    \vspace{-4mm} %
\end{figure*}

We propose the \textbf{Caption--Prompt--Judge (CPJ)} framework, a training-free pipeline leveraging LLM-as-a-Judge self-refined captions for debiased, evidence-grounded Agri-Pest VQA. As illustrated in Fig. \ref{fig:framework}, CPJ operates through: (i) generative interpretable captioning with iterative refinement via LLM-as-a-Judge, (ii) caption-informed VQA using task-specific prompts, and (iii) answer selection based on diagnostic validity and practical relevance.

Evaluated on CDDMBench~\cite{liu2024multimodal} using lightweight LVLMs (Qwen-VL-Chat, GPT-5-Nano), CPJ achieves significant improvements via strict keyword matching for classification and LLM-based scoring for QA. GPT-5-Nano with GPT-5-mini captions achieves \textbf{+22.7} pp in disease classification and \textbf{+19.5} points in QA score over baselines, demonstrating enhanced accuracy and transparency without supervised fine-tuning.

Our contributions include:
\begin{itemize}
\item \textbf{Framework}: A training-free CPJ framework prioritizing interpretability through iteratively refined captions, enabling transparent diagnostic reasoning and addressing the gap between black-box predictions and explainable AI needs.
\item \textbf{Methodology}: Integration of multi-angle descriptions with LLM-as-a-Judge refinement and dual-answer VQA generation, bridging visual-semantic gaps while maintaining objectivity through crop/disease name exclusion in caption prompts.
\item \textbf{Empirical Validation}: Comprehensive evaluation on CDDMBench showing significant gains: GPT-5-Nano achieves +22.7pp in disease classification and +19.5 points in QA score with GPT-5-mini captions, with human expert validation (94.2\% agreement, Cohen's $k = 0.88$) confirming LLM-as-a-Judge reliability.
\end{itemize}

\section{RELATED WORK}
\label{sec:related}

\subsection{Agricultural Vision-Language Models}
Recent advancements in Vision-Language Models (VLMs) have enabled domain-specific adaptations for agricultural disease diagnosis. Traditional approaches to crop disease identification relied on handcrafted features and classical machine learning~\cite{kumar2015plant}, but the advent of deep learning, particularly Vision Transformers (ViT)~\cite{dosovitskiy2020image}, has significantly improved classification accuracy. Pre-trained language models have also shown promise in agricultural text classification tasks~\cite{yadav2023comparative}. Zhu et al.~\cite{zhu2024harnessinglargevisionlanguage} comprehensively review VLM applications in agriculture, covering pest detection, weed identification, and seed quality assessment, while addressing challenges such as data scarcity, domain shift, and training efficiency. Specialized models like AgroGPT and AgriGPT-VL~\cite{yang2025agrigptvlagriculturalvisionlanguageunderstanding} demonstrate the efficacy of domain-specific training: Yang et al. train AgriGPT-VL on the large-scale Agri-3M-VL dataset via a progressive curriculum from text grounding to multimodal alignment, achieving state-of-the-art performance on agricultural benchmarks. Similarly, AgriVLM~\cite{yu2024framework} integrates ViT and ChatGLM via Q-Former with LoRA fine-tuning, achieving over 90\% accuracy in crop growth stage and disease recognition through parameter-efficient adaptation. Few-shot learning approaches~\cite{zhou2024few} further enhance model adaptability in low-resource scenarios, while specialized agricultural tasks benefit from vision language model leveraging~\cite{arshad2025leveraging}.

Benchmark evaluations reveal significant challenges in agricultural VQA~\cite{su2022survey}. Shinoda et al.~\cite{Shinoda_2025_ICCV} introduce AgroBench, an expert-annotated benchmark covering 203 crop and 682 disease categories across seven tasks, revealing that over 50\% of errors stem from lack of domain-specific knowledge. Ranario and Earles~\cite{ranario2025vision} benchmark 27 agricultural classification tasks, finding that zero-shot VLMs underperform supervised baselines and that multiple-choice prompting significantly outperforms open-ended generation. 
While large-scale datasets (e.g., MIRAGE~\cite{dongre2025mirage}, CDDMBench~\cite{liu2024multimodal},AgMMU~\cite{gauba2025agmmu} ) enable supervised fine-tuning (SFT) methods to achieve strong in-domain performance~\cite{yang2024application}, these approaches tend to overfit narrow distributions and lack transparent reasoning. This limits their robustness to domain shifts (e.g., geography, season) and scalability in resource-limited agricultural settings~\cite{saadati2024out}.

\subsection{Explainable AI in Vision-Language Models}
The opacity of VLMs necessitates explainability frameworks to enhance transparency and trustworthiness across diverse domains. Dinga et al. employ Shapley Values to interpret VLM-generated map descriptions, quantifying region-specific contributions to textual outputs and revealing how map labels and scale influence caption accuracy~\cite{dinga2025you}. In medical imaging, Kamal et al. integrate U-Net, Vision Transformer, and Medical Language Graphs with LRP and LIME to interpret image captioning models, improving transparency in diagnostic report generation~\cite{kamal2025explainable}. Masala and Leordeanu introduce GEST, a spatio-temporal event graph framework for explainable video description, converting structured graphs into natural language to enhance interpretability in vision-to-language tasks~\cite{masala2025vision}. 
Natarajan and Nambiar's VALE~\cite{natarajan2024vale} combines SHAP, SAM, and VLMs to generate visual-textual rationales for image classification via prompt engineering. Saakyan et al.~\cite{saakyan2025understanding} investigate VLMs' figurative meaning comprehension via the V-FLUTE dataset, framing it as explainable visual entailment. Park et al.~\cite{park2025leveraging} explore MLLMs for interface understanding, demonstrating cross-domain vision-language reasoning.

The LLM-as-a-Judge paradigm~\cite{li2024llms} has emerged as a powerful evaluation mechanism across diverse AI tasks, including code understanding and translation quality assessment. Our CPJ framework extends these XAI and evaluation principles to agricultural diagnosis, employing structured captions and LLM-based refinement to enhance interpretability without domain-specific annotations.

\section{METHODOLOGY}
\label{sec}
\subsection{Image Interpretability and Caption Enhancement}
\label{subsec:caption_enhancement}
\textbf{Motivation.} Raw agricultural images often contain subtle visual cues that are difficult for models to interpret directly—such as early-stage lesions, minor discoloration, or texture variations indicative of nutrient deficiencies. Existing vision-language models, while powerful in open-domain tasks, may overlook these domain-specific details when processing images end-to-end. By explicitly generating detailed, structured captions as an intermediate step, we provide the model with a richer semantic context that highlights critical diagnostic features, thereby improving both accuracy and interpretability.

To achieve high-quality visual question answering (VQA) in agriculture, we enhance semantic descriptions of input crop images using multimodal large models without fine-tuning. Leveraging few-shot prompting, our method generates detailed captions including morphological traits (leaf shape, texture, size), symptom characteristics (lesion patterns, discoloration, necrosis), and developmental stages. These captions serve as rich, interpretable intermediate representations that bridge the semantic gap between raw pixel data and agricultural domain knowledge.

\textbf{Bias Mitigation.} To minimize bias in subsequent VQA, prompts explicitly exclude crop or disease names, focusing on objective morphological and symptomatic features. This prevents premature diagnostic conclusions and ensures captions remain neutral observations.

The initial caption, generated via a vision-language model (VLM) with few-shot prompting, is:
\begin{equation}
\vspace{-1.0mm}
C_0 = \mathcal{M}_{\text{VLM}}(I, P_{\text{few}}),
\label{eq:caption}
\vspace{-1.0mm}
\end{equation}
where $I$ is the image and $P_{\text{few}}$ is the prompt template.

A large language model (LLM) evaluates caption quality $s(C_0)$ based on \textbf{accuracy, completeness, and neutrality}~\cite{zheng2023judging}. Captions below the threshold $\tau=8.0$ are refined:
\begin{equation}
\vspace{-1.0mm}
C^\ast =
\begin{cases}
C_0, & s(C_0) \geq \tau, \\
\mathcal{M}_{\text{VLM}}(I, \mathcal{R}(C_0)), & \text{otherwise},
\end{cases}
\label{eq:refinement}
\vspace{-1.0mm}
\end{equation}
where $\mathcal{R}(\cdot)$ provides targeted refinement instructions. This iterative process ensures captions are interpretable and semantically dense, forming a robust foundation for VQA tasks.

\subsection{Explanational caption-Optimized VQA}
\label{subsec}

Using refined captions from stage one, we implement a multimodal VQA pipeline \cite{wang2024agri} for crop disease classification and knowledge answering. Given input $X = (I, C^\ast, Q)$, where $Q$ is the query, the VQA model generates two complementary answers:
\begin{equation}
\vspace{-1.0mm}
\mathcal{A} = \{A^{(1)}, A^{(2)}\} = \mathcal{M}_{\text{VQA}}(X, P_{\text{task}}),
\label{eq:vqa}
\vspace{-1.0mm}
\end{equation}
where $P_{\text{task}}$ is the task-specific prompt template.

\subsubsection{Crop Disease Classification Task}
\label{subsubsec}

This task identifies crop species and disease type through joint analysis of visual and textual data. Two complementary outputs ensure comprehensive recognition: 1). Pest/disease identification: Symptoms, severity, and characteristic features of the plant and disease.
2). Crop identification: Specific crop type, variety, and distinctive morphological traits.

\begin{table*}[!t]
    \renewcommand{\arraystretch}{0.82}
    \caption{Performance comparison of Qwen-VL-Chat and GPT-5-Nano across two explanatory captions generated by Qwen2.5-VL-72B-Instruct and GPT-5-mini on the CDDMBench dataset for two tasks. Strategies include ``Explanational Caption" (optimized captions), ``Explanational Caption + few-shot" (with in-context examples), and ``+LLM-as-a-judge" (selecting the best of two answers per question). Scores represent the highest of two candidates, with best results in \textbf{bold} for each type of model.}
    \resizebox{\textwidth}{!}{
    \renewcommand\tabcolsep{4pt}
        \begin{tabular}{lccc}\toprule
        \multirow{2}{*}{\textbf{Model}} & \multicolumn{2}{c}{\textbf{Crop Disease Diagnosis}} & \multirow{2}{*}{\textbf{Crop Disease Knowledge QA}} \\
        \cmidrule(lr){2-3}
        & \textbf{Crop Classification} & \textbf{Disease Classification} & \\
        \midrule
        \multicolumn{4}{c}{\textbf{Generated Explanation Caption (Qwen2.5-VL-72B-Instruct)}} \\
       \midrule
        Qwen-VL-Chat (Liu et al. (2024) Baseline) & 28.40\% & 5.00\% & 41 \\
        Qwen-VL-Chat (Our Baseline) & 28.55\% & 5.80\% & 41.5 \\
        Qwen-VL-Chat (Explanational Caption) & 29.30\% & 12.10\% & 46.5 \\
        Qwen-VL-Chat (~~~~+few shot) & 53.39\% & 24.49\% & 50 \\
        \textbf{Qwen-VL-Chat (~~~~~~+llm-as-a-judge)} & \textbf{54.90\%} & \textbf{25.39\%} & \textbf{51} \\
        \cdashline{1-4}
        Gpt-5-Nano (Our Baseline) & 47.00\% & 11.00\% & 65 \\
        Gpt-5-Nano (Explanational Caption) & 51.20\% & 31.40\% & 75.5 \\
        Gpt-5-Nano (~~~~+few-shot) & 50.10\% & 31.00\% & 74.5 \\
        \textbf{Gpt-5-Nano (~~~~~~~~+llm-as-a-judge)} & \textbf{53.20\%} & \textbf{32.80\%} & \textbf{76} \\
        \midrule
        \multicolumn{4}{c}{\textbf{Generated Explanation Caption (GPT-5-mini)}} \\
       \midrule
        Qwen-VL-Chat (Explanational Caption) & 28.99\% & 7.17\% & 44 \\
        Qwen-VL-Chat (~~~~+few-shot) & 51.00\% & 14.50\% & 49 \\
        \textbf{Qwen-VL-Chat (~~~~~~~~+llm-as-a-judge)} & \textbf{51.60\%} & \textbf{14.80\%} & \textbf{49.5} \\
        \cdashline{1-4}
        Gpt-5-Nano (Our Baseline)& 47.00\% & 11.00\% & 65 \\
        Gpt-5-Nano (Explanational Caption) & 60.30\% & 31.60\% & 84 \\
        Gpt-5-Nano (~~~~+few-shot) & 58.90\% & 29.80\% & 76 \\
        \textbf{Gpt-5-Nano (~~~~~~~~+llm-as-a-judge)} & \textbf{63.38\%} & \textbf{33.70\%} & \textbf{84.5} \\
        \bottomrule
        \end{tabular}    
        }
\label{tab:crop-disease-results}
\end{table*}

\subsubsection{Crop Disease Knowledge Question Answering Task}
\label{subsubsec}

This task generates actionable agricultural knowledge with expert-level prompts \cite{gauba2025agmmu}. Outputs include:  1). Treatment, prevention, and control recommendations with specific methods.
2). Disease explanation: Symptoms, etiology, and lifecycle mechanisms.

Prompts ensure \textbf{two complementary answers} per query, addressing both recognition and explanation. Few-shot examples include images, explanational captions, questions, and the two-answer structure to maintain contextual consistency. Prompt engineering supports zero- to five-shot settings, ensuring scalability and efficiency.

\subsection{LLM-as-a-Judge Answer Selection}
\label{subsec:llm_judge}

To enhance robustness and reduce uncertainty in VQA outputs, we introduce an LLM-as-a-Judge stage that evaluates and selects the most reliable answer from dual-answer candidates, addressing hallucination mitigation and completeness verification.

A stronger LLM (e.g., GPT-4) evaluates both VQA answers using multi-dimensional criteria $\Omega$. For the disease diagnosis task, criteria include: (i) \textit{plant accuracy} (correct crop species identification), (ii) \textit{disease accuracy} (correct disease/pest identification), (iii) \textit{symptom accuracy} (precise symptom description), (iv) \textit{format adherence} (both plant and disease information present), and (v) \textit{completeness} (comprehensive and professional response). For the knowledge QA task, criteria encompass: (i) \textit{accuracy} (scientifically correct information), (ii) \textit{completeness} (covers all relevant aspects), (iii) \textit{specificity} (precise details on rates, timings, methods), (iv) \textit{practicality} (actionable for farmers), and (v) \textit{scientific validity} (evidence-based with proper terminology). Each criterion is scored, and the total score guides selection:
\begin{equation}
\vspace{-1.0mm}
\text{Score}(A) = \frac{1}{|\Omega|} \sum_{\omega \in \Omega} g_{\omega}(A, A_{\text{ref}}),
\label{eq:judge}
\vspace{-1.0mm}
\end{equation}
where $g_\omega(\cdot)$ is the scoring function for criterion $\omega$, and $A_{\text{ref}}$ is the reference answer.

The answer with the highest score is selected:
\begin{equation}
\vspace{-1.0mm}
A^\ast = \arg\max_{A \in \mathcal{A}} \text{Score}(A).
\label{eq:selection}
\vspace{-1.0mm}
\end{equation}

The LLM provides an evaluation report detailing strengths and weaknesses of each answer, key differences from the reference, and justifications for scoring decisions, ensuring diagnostic transparency. The system outputs $A^\ast$ and the evaluation report, delivering reliable recommendations.

\section{Experiments and Analysis}
\label{sec}

\subsection{Dataset and Evaluation}
We evaluate our framework on the CDDMBench dataset~\cite{liu2024multimodal} across two tasks, following the original evaluation settings, metrics, and test splits.

For the \textbf{Crop Disease Diagnosis} task, the test set includes 3,000 images withheld from training. Each explanatory question pair assesses the model's ability to identify crop and disease types, with performance measured by keyword matching in responses.

For the \textbf{Crop Disease Knowledge QA} task, the test set contains 20 questions spanning 10 crop disease types, paired with images. Responses are scored by GPT-4 on usefulness, relevance, and accuracy (1–10), with scores normalized to a 100-point scale. All evaluation protocols strictly adhere to~\cite{liu2024multimodal}.

\subsection{Model Selection and Settings}
To enhance interpretability, we employ two large vision-language models, Qwen2.5-VL-72B-Instruct and GPT-5-mini, to generate explanatory descriptive captions for each input image. These captions describe plant type, disease symptoms, severity, and uncertainties. Each caption is evaluated by the same model acting as an LLM-as-a-Judge, and those scoring below 8 out of 10 are refined iteratively. For VQA, we use the 7B Qwen-VL-Chat model and GPT-5-Nano under few-shot settings, with key parameters set as: \textit{Qwen-VL-Chat} (temperature: 0.5, max\_tokens: 400, top\_p: 0.8, max\_retries: 3) and \textit{GPT-5-Nano} (reasoning\_effort: medium, verbosity: low, max\_retries: 3, timeout: 30).
%including temperature=$0.5$, max\_tokens=$400$ for Qwen, and reasoning\_effort=medium for GPT-5-Nano.
The final answer is selected by GPT-5 as LLM-as-a-Judge Answer Selection based on accuracy and specificity without image access to avoid bias. Our framework is implemented using LangChain~\cite{annam2025langchain,topsakal2023creating,dave2025learning} for prompt management, model execution, and pipeline design. All models are accessed via API.

\subsection{Main Results}
\label{subsec:main-results}

Table~\ref{tab:crop-disease-results} compares Qwen-VL-Chat and GPT-5-Nano on Crop Disease Diagnosis and Crop Disease Knowledge QA tasks across various caption strategies. The combination of Explanational Caption, few-shot prompting, and LLM-as-a-Judge delivered the best performance for both models.

\textbf{Overall Performance.} Using GPT-5-mini captions, GPT-5-Nano achieved best results at 63.38\%, 33.70\%, and 84.5, representing gains of +22.7pp in disease classification and +19.5 points in QA over baselines. With Qwen2.5-VL-72B captions, Qwen-VL-Chat scored 54.90\%, 25.39\%, and 51.0, while GPT-5-Nano scored 53.20\%, 32.80\%, and 76.0.

\textbf{Model-Specific Patterns.} GPT-5-Nano benefits more from caption quality while Qwen-VL-Chat exhibits stronger in-context learning from few-shot prompting. This suggests reasoning-focused models rely more on caption quality for grounding, while vision-centric models benefit more from example-based learning, with ablation results in Sec.~\ref{subsec:ablation}.

\textbf{Cross-Caption Generator Analysis.} GPT-5-mini captions outperform Qwen2.5-VL-72B captions (+9.08pp crop, +2.10pp disease), indicating caption generation capability directly influences downstream performance. Few-shot prompting shows differential effects: Qwen-VL-Chat gains substantially (+24.10pp crop), while GPT-5-Nano shows minor decreases (-1.40pp crop), suggesting vision-centric models benefit more from visual patterns while reasoning models leverage captions effectively in zero-shot settings.

\begin{figure}[!t]
   % \vspace{-0.5em} % 缩小顶部间距
    \centering
    \includegraphics[width=0.96\linewidth]{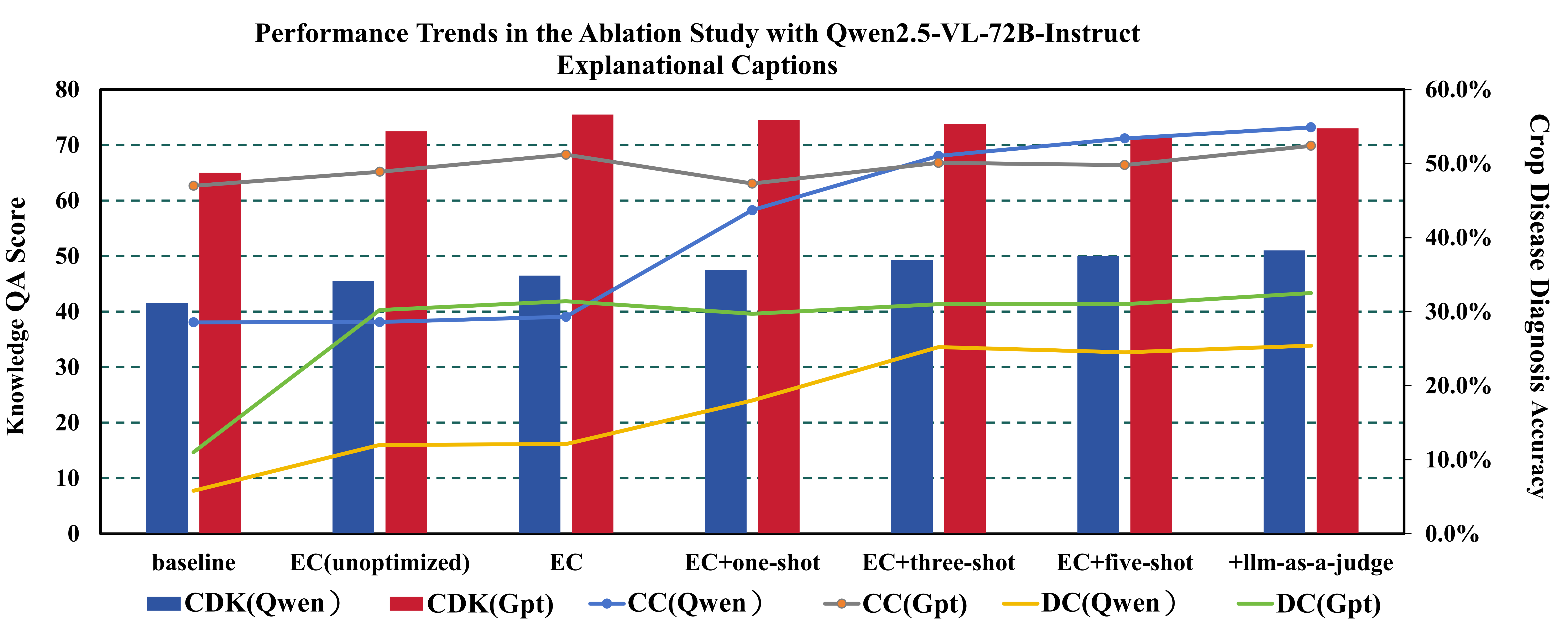} % 换成你的文件名，修改高度
    \vspace{-2mm} % 缩小底部间距
    \caption{
Performance trends in the ablation study using Qwen2.5-VL-72B-Instruct with Explanational Captions (EC). The abbreviations are as follows: EC for Explanational Caption, CC for Crop Classification, DC for Disease Classification, CDK for Crop Disease Knowledge, Qwen for Qwen-VL-Chat, and GPT for GPT-5-Nano.
}
    \label{fig:ablation-study}
    \vspace{-3mm}
\end{figure}

%\section{Analysis}
\subsection{Ablation Study}
\label{subsec:ablation}

This section examines component-wise contributions from Sec.~\ref{subsec:main-results}. As shown in Fig.~\ref{fig:ablation-study}, both Qwen-VL-Chat and GPT-5-Nano steadily benefit from optimized captions, few-shot prompting, and LLM-as-a-Judge.
We perform ablations under the \textit{Qwen2.5-VL-72B-Instruct} caption setting to quantify each module's role.

\textbf{Baseline $\rightarrow$ Unoptimized Captions.}
For Qwen-VL-Chat, unoptimized captions slightly improve Crop Classification (+0.05\,pp) but significantly enhance Disease Classification (+6.20\,pp) and QA (+4.0).
GPT-5-Nano shows larger improvements (+1.90\,pp / +19.20\,pp / +7.5), indicating higher sensitivity to contextual information.

\textbf{Caption Optimization.}
Iterative refinement boosts Qwen-VL-Chat by +0.70\,pp / +0.10\,pp / +1.0 and GPT-5-Nano by +2.30\,pp / +1.20\,pp / +3.0, demonstrating benefits from higher semantic density and reduced label leakage.

\textbf{Few-shot Prompting.}
Qwen-VL-Chat achieves the largest boost at three-shot (+22.50\,pp / +19.40\,pp / +7.8), showing strong in-context learning capabilities. GPT-5-Nano shows smaller changes, suggesting robust baseline reasoning.

\textbf{Five-shot + LLM-as-a-Judge.}
Best results: Qwen-VL-Chat reaches 54.90\%, 25.39\%, and 51.0; GPT-5-Nano achieves 52.40\%, 32.50\%, and 73.0. The LLM-as-a-Judge stage effectively filters hallucinations and selects factually complete answers, enhancing diagnostic reliability.

\subsection{Qualitative Analysis and Error Patterns}
\label{subsec:qualitative}

To understand CPJ's performance mechanisms, we analyze 150 sampled predictions. Table~\ref{tab:error_patterns} categorizes behaviors into success cases and common error patterns.

\begin{table}[!t]
    \centering
    \caption{Error pattern analysis with frequency distribution from 150 sampled predictions. The table presents success cases and four primary error categories: early-stage ambiguous symptoms (18\%), co-infection confusion (12\%), low caption quality (8\%), and judge bias toward verbosity (5\%).
    }
    \label{tab:error_patterns}
    \small
    \begin{tabular}{p{2.5cm}p{5.2cm}}
        \toprule
        \textbf{Category} & \textbf{Description \& Frequency} \\
        \midrule
        Success Case & Bacterial leaf spot: Caption ``necrotic lesions, yellow halos'' → correct diagnosis (9/10 QA) \\
        \midrule
        Error:Ambiguous & Early-stage subtle symptoms ($\sim$18\%) \\
        Error:Co-infection & Mixed diseases confuse prediction ($\sim$12\%) \\
        Error:Caption quality & Low-quality captions degrade VQA ($\sim$8\%) \\
        Error:Judge bias & Favors verbose responses ($\sim$5\%) \\
        \bottomrule
    \end{tabular}
    \vspace{-2mm}
\end{table}

\textbf{Success mechanisms.} Explanational Captions enable precise pathogen identification by translating visual symptoms into structured semantic features. For bacterial leaf spot, caption descriptions such as ``necrotic lesions, yellow halos'' directly map to pathogen-specific diagnostic criteria, enabling accurate treatment protocols. LLM-as-a-Judge shows strong factuality verification (73\% vs. 51\% correctness), effectively filtering hallucinations through multi-criteria scoring.

\textbf{Failure modes.} The dominant error (18\%) stems from early disease stage ambiguity, where minimal visual symptoms provide insufficient diagnostic evidence even after caption enhancement. Co-infection scenarios (12\%) require multi-label architectures. Caption quality issues (8\%) correlate with image artifacts (blur, occlusion, lighting), suggesting adaptive preprocessing. Judge bias (5\%) indicates need for length-normalized scoring.

\section{Conclusion and Limitations}

We proposed a parameter-free, environment-adaptive generate--judge--select pipeline for agricultural VQA, coupling debiased, optimized captions with a two-answer VQA design and an LLM-as-a-Judge selector. On CDDMBench, the framework improves recognition and knowledge QA without SFT: GPT-5-mini captions help GPT-5-Nano achieve 63.38\% (Crop), 33.70\% (Disease), and 84.5 (QA); Qwen-VL-Chat attains 54.90\%, 25.39\%, and 51.0, outperforming caption-free baselines. The method is lightweight, reproducible, and enhances interpretability through structured evidence and rubric-based judgments. By decomposing the diagnostic process into explicit captioning, dual-answer generation, and meta-evaluation stages, CPJ provides practitioners with transparent reasoning chains that connect visual observations to actionable recommendations, fostering trust and enabling informed decision-making in AI-driven agricultural decision support systems across diverse farming contexts.

Limitations include sensitivity to caption quality, judge bias toward verbose responses, and lack of temporal/multi-view context. Caption refinement significantly mitigates quality issues, yet edge cases with severe image degradation remain challenging. The current framework operates on single images, limiting its ability to capture disease progression or environmental variations across time and spatial scales. Future work will explore knowledge-grounded verification, uncertainty calibration, multi-image and multilingual settings, cross-regional adaptation, and on-device deployment for field applications in resource-constrained environments.

%\newpage
\bibliographystyle{IEEE_Conference_Template-ICME_2026/IEEEbib}
\bibliography{IEEE_Conference_Template-ICME_2026/icme2026references}

\end{document}